\documentclass[letterpaper, 10 pt, conference]{ieeeconf}

\IEEEoverridecommandlockouts                              

\usepackage{amsfonts}
\usepackage{amsmath}
\usepackage{cite}
\usepackage{graphicx}
\usepackage{float} 
\usepackage{subfigure}
\usepackage{overpic}
\usepackage{wrapfig}
\usepackage{footmisc}
\usepackage{amssymb}

\title{\LARGE \bf
Improving Neural Indoor Surface Reconstruction with Mask-Guided Adaptive Consistency Constraints
}

\author{Xinyi Yu$^{1}$, Liqin Lu$^{1}$, Jintao Rong$^{1}$, Guangkai Xu$^{2,*}$ and Linlin Ou$^{1}$ \ 
\thanks{$^{1}$Xinyi Yu, Liqin Lu, Jintao Rong, and Linlin Ou are with  College of Information Engineering, Zhejiang University of Technology, Hangzhou, China.{\tt\small \{yuxy,201806060214,2111903071,linlinou\}@zj\ ut.edu.cn}} \
\thanks{$^{2}$Guangkai Xu is with College of Computer Science and Technology, Zhejiang University, Hangzhou, China. {\tt\small guangkai.xu@zju.edu.cn }} \
\thanks{$^{*}$ Corresponding author.}
}

\begin{document}

\maketitle
\thispagestyle{empty}
\pagestyle{empty}

\begin{abstract}

3D scene reconstruction from 2D images has been a long-standing task. Instead of estimating per-frame depth maps and fusing them in 3D, recent research leverages the neural implicit surface as a unified representation for 3D reconstruction. Equipped with data-driven pre-trained geometric cues, these methods have demonstrated promising performance. However, inaccurate prior estimation, which is usually inevitable, can lead to suboptimal reconstruction quality, particularly in some geometrically complex regions. In this paper, we propose a two-stage training process, decouple view-dependent and view-independent colors, and leverage two novel consistency constraints to enhance detail reconstruction performance without requiring extra priors. Additionally, we introduce an essential mask scheme to adaptively influence the selection of supervision constraints, thereby improving performance in a self-supervised paradigm. Experiments on synthetic and real-world datasets show the capability of reducing the interference from prior estimation errors and achieving high-quality scene reconstruction with rich geometric details.


\end{abstract}


\section{Introduction}
3D scene reconstruction from multiple images is a fundamental vision task with diverse applications, including robotics~\cite{sajjan2020clear, yang2021robotic, skinner2017automatic}, virtual reality, augmented reality, etc. In robotics, reconstructions are used in trajectory planning~\cite{yang2021robotic} and mapping~\cite{wang2023real}. Given posed images, traditional algorithms usually estimate depth maps and lift them into 3D space, which can be categorized into multi-view stereo methods and monocular depth estimation methods. Multi-view stereo (MVS~\cite{schonberger2016pixelwise, schonberger2016structure, seitz2006comparison}) leverages accurate feature correspondences between keyframes to recover the 3D structure, and monocular depth estimation relies on large-scale training datasets to improve the generalization to diverse scenes. While feature matching lacks confidence in lighting changes, occlusion, and low-texture regions, and robust monocular depth estimation usually suffers from the unknown scale, their performance can hardly deal with multi-frame inconsistency and is less satisfactory. Although some RGB-D fusion algorithms~\cite{whelan2015real, wang2022rgb} and post-processing optimization methods~\cite{Xu_2023_ICCV, bian2021unsupervised, bian2021auto} are committed to ensuring
consistency, it is found that they have difficulty handling some inaccurate depth predictions.


In order to tackle the consistency problem, there is an urgent need for a unified 3D representation instead of per-frame 2D depth maps. Some learning-based methods~\cite{murez2020atlas, sun2021neuralrecon} project 2D features to spatial and directly predict TSDF value in 3D position. On the other hand, armed with the volume rendering theory, optimization-based methods usually encode a specific scene with a neural implicit scene representation by overfitting the pair of the input 3D position and the output color and geometry. However, unlike object-centric cases, the sparse views and texture-less areas of indoor scenes may lead to limited surface quality and local minima in optimization. To address the issue, some approaches integrate affine-invariant depth~\cite{xu2022towards, yin2021learning, ranftl2020towards, roessle2022dense, neff2021donerf} and predicted normal priors~\cite{yu2022monosdf, wang2022neuris} as supervision. Although promising results have been achieved, they struggle to handle both the unknown scale-shift values of depth priors and inaccurate prior estimations, which results in poor reconstruction quality of complex geometric details.


In this study, we propose mask-guided adaptive consistency constraints to improve the detail reconstruction performance of neural surface representation. Similar to previous works~\cite{yu2022monosdf, wang2022neuris, yariv2020multiview}, we also use two kinds of MLPs to predict signed distance and color information and employ the surface normal predicted by a pre-trained model~\cite{eftekhar2021omnidata} as priors. 
Specifically, we divide our training process into two stages. In the first stage, we focus on optimizing the RGB image reconstruction constraint as our primary objective while incorporating the estimated normal vector cues as additional supervision signals. This allows us to obtain an initial scene geometry. In the second stage, based on the principle that excellent reconstruction quality is closely associated with multi-view consistency, we categorize the reconstructed components into two groups: the accurate part and the inaccurate part, based on the difference in the rendered normal vectors at different viewpoints.
During training, except for the color reconstruction constraint, to the sampled rays passing through the inaccurate part, we employ a geometric consistency constraint to improve the accuracy of rendered depth; For the accurate part, we continue to apply normal priors to supervision. Besides, we decouple the color map into a view-dependent one and a view-independent one, and leverage another photometric consistency constraint to supervise the view-independent color. 
This two-stage training strategy adaptively distinguishes the accuracy of surface normal priors and adopts different supervision paradigms separately.
Experimental results on both synthetic and real-world datasets demonstrate that our method achieves high-quality reconstructions with rich geometric details, outperforming other existing methods. Our main contributions are summarized as follows:

\begin{itemize}
  \item[$\bullet$]We propose a two-stage training process for neural surface reconstruction, which decouples the view-dependent and view-independent colors and leverages two novel consistency constraints to improve the quality.
  \item[$\bullet$]
  Based on the essential mask scheme, our model can reduce the side effects of inaccurate surface normal priors and enhance performance in a self-supervised paradigm during training.
  \item[$\bullet$]
  Experimental results on both synthetic and real-world datasets show that we can achieve high-quality scene reconstruction performance with rich geometric details.
\end{itemize}


\section{related work}
\subsection{Multi View Stereo}
For years, 3D reconstruction from multi-view perspectives has remained a challenging yet significant task. Multi-view stereo (MVS) is a traditional reconstruction approach that leverages feature matching and triangulation methods to estimate 3D positions corresponding to pixels or features across multi-views. Using estimated positions and bundle adjustment, MVS recovered depth or normal to recover geometry~\cite{seitz2006comparison, schonberger2016pixelwise, furukawa2015multi, lindenberger2021pixel}. While MVS has achieved significant success, the estimated depth suffers from inaccurate and scale-inconsistent in abundant specular reflection or untextured regions, such as indoor scenes. Although some RGB-D fusion~\cite{whelan2015real, wang2022rgb} and post-processing optimization methods~\cite{Xu_2023_ICCV, bian2021unsupervised, bian2021auto} have handled scale consistency issues, they still fail in inaccurate estimations. With the advancements in deep learning, learning-based methods have witnessed significant development in recent years. These methods employ neural networks to estimate depth or TSDF (truncated signed distance function) end-to-end. Depth-based methods estimate depth maps and utilize fusion procedures to reconstruct, which encounter challenges in noisy surfaces, and scale ambiguities. TSDF-based methods like NeuralRecon~\cite{sun2021neuralrecon} propose a novel framework to lift the 2D features, fuse them spatially and temporally in 3D space, and predict the TSDF volume directly. Those methods always produce overly smooth reconstructions.

\subsection{Implicit Representation of Geometry}
Recently, with the success of volume rendering theory, some methods leverage Multi-Layer Perceptrons (MLPs) to implicitly represent geometry. These approaches are supervised by RGB images, overfit the pair of the 3D coordinates, and the corresponding color and geometric properties, like volume density or occupancy~\cite{mildenhall2021nerf, zhu2022nice, tewari2020state, mescheder2019occupancy, wang2021neus}. One notable advancement in this area is the Neural Radiance Field (NeRF) technique, which has demonstrated remarkable results in both novel view synthesis~\cite{barron2022mip, johari2022geonerf} and implicit surface reconstruction~\cite{wang2021neus, yariv2020multiview, yariv2021volume}. These approaches utilize volume rendering methods to supervise implicit scene representation through a 2D photometric loss. However, volume density representations often fall short in geometric details due to a lack of sufficient constraints. Some methods have attempted to improve the volume rendering framework, such as VolSDF~\cite{yariv2021volume} and NeuS~\cite{wang2021neus}, which have achieved better surface reconstructions but still face challenges in reconstructing geometric details. Consequently, certain methods aim to integrate geometric cues acquired from sensors~\cite{yan2023efficient, azinovic2022neural} or predicted by models~\cite{yu2022monosdf, wang2022neuris, wang2022neuralroom} to strengthen geometric constraints. MonoSDF\cite{yu2022monosdf} integrates estimated monocular geometric clues into a neural volume rendering framework to enhance the overall quality of the reconstruction. NeuRIS~\cite{wang2022neuris} adaptively utilizes predicted normal cues by patch matching between neighboring images. While these methods have achieved accurate reconstruction results, they are sensitive to the accuracy in priors, especially when applied to real scenes. Compared to these methods, our method utilizes normal vector cues more efficiently and performs better in real-world scenarios.


\section{Method}

\begin{figure*}[h]
\centering
\includegraphics[width=0.94\textwidth]{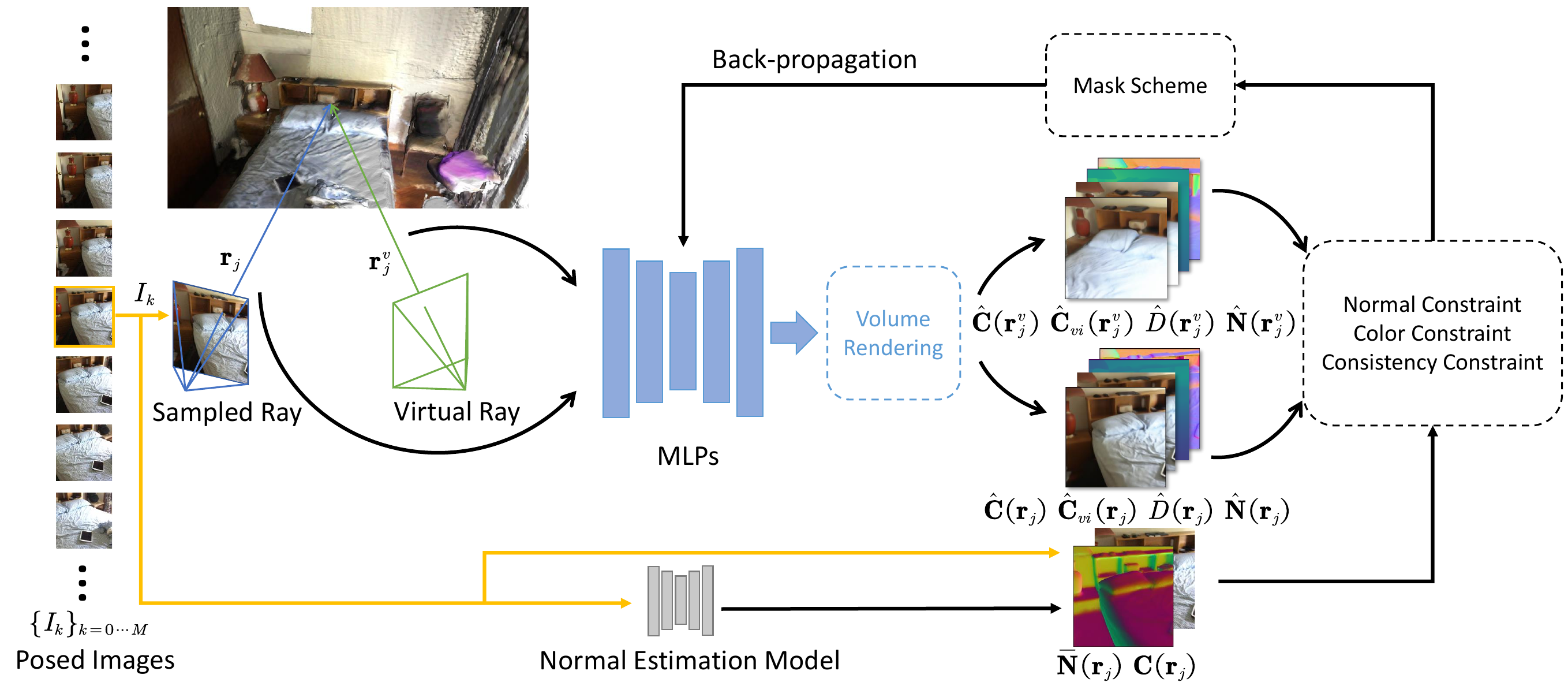}


\caption{We sample rays pass through selected image $I_k$. Randomly generating virtual ray $\mathbf{r}^v_j$ corresponding to each sample ray $\mathbf{r}_j$. Rendering the color $\hat{\mathbf{C}}$, view-independent color $\hat{\mathbf{C}}_{vi}$, depth $\hat{D}$ and normal $\hat{\mathbf{N}}$ along these rays. Using a pre-trained model to estimate normal priors $\bar{\mathbf{N}}$ of $\mathbf{r}_j$. To learn MLPs' weights, we minimize the difference between $\hat{\mathbf{C}}(\mathbf{r}_j)$ and the given color $\mathbf{C}(\mathbf{r}_j)$. Besides, we utilize the mask-guided consistency and normal constraints.}

\label{pipline}
\vspace{-0.4cm}
\end{figure*}

Aiming at 3D indoor scene reconstruction from posed images $\{I_k\}_{k=0\cdots M}$, we use a neural surface representation optimized through the supervision of rendered RGB images. To enhance the robustness, normal priors obtained from pre-trained models are adopted. However, directly supervising the rendered normal may disturb the training process if the normal priors are inaccurate. Therefore, We optimize it with color and normal constraints to get the initial shape in the first stage. Then, we introduce geometric and photometric constraints to further improve reconstruction quality. What's more, all the training constraints except the color one are guided by our proposed mask scheme in stage two. The overall pipeline of the second stage is shown in Fig.~\ref{pipline}.

Concretely, for each selected image $I_k$, we sample $m$ rays $\{\mathbf{r}_j\}_{j=0\cdots m}$ passing though it. Randomly generating a virtual ray $\mathbf{r}^v_j$ to pair with the current sampled ray $\mathbf{r}_j$. Using MLPs and volume rendering framework, we render color $\hat{\mathbf{C}}$, decomposed color $\hat{\mathbf{C}}_{vi}$, depth $\hat{D}$, and normal $\hat{\mathbf{N}}$ for both $\mathbf{r}_j$ and $\mathbf{r}_j^v$. To train our model, we minimize the color and normal differences, between $\hat{\mathbf{C}}(r_j)$ and the given color $\mathbf{C}(\mathbf{r}_j)$, $\hat{\mathbf{N}}(\mathbf{r}_j)$ and the normal prior $\bar{\mathbf{N}}(\mathbf{r}_j)$ predicted by pre-trained models~\cite{eftekhar2021omnidata}, respectively. Furthermore, we introduce a multi-view consistency constraint to guide the training process. During optimization, we use a masking scheme to adaptively select different training constraints for different sampled rays.


\begin{figure}[htp]
\centering
\includegraphics[width=0.45\textwidth]{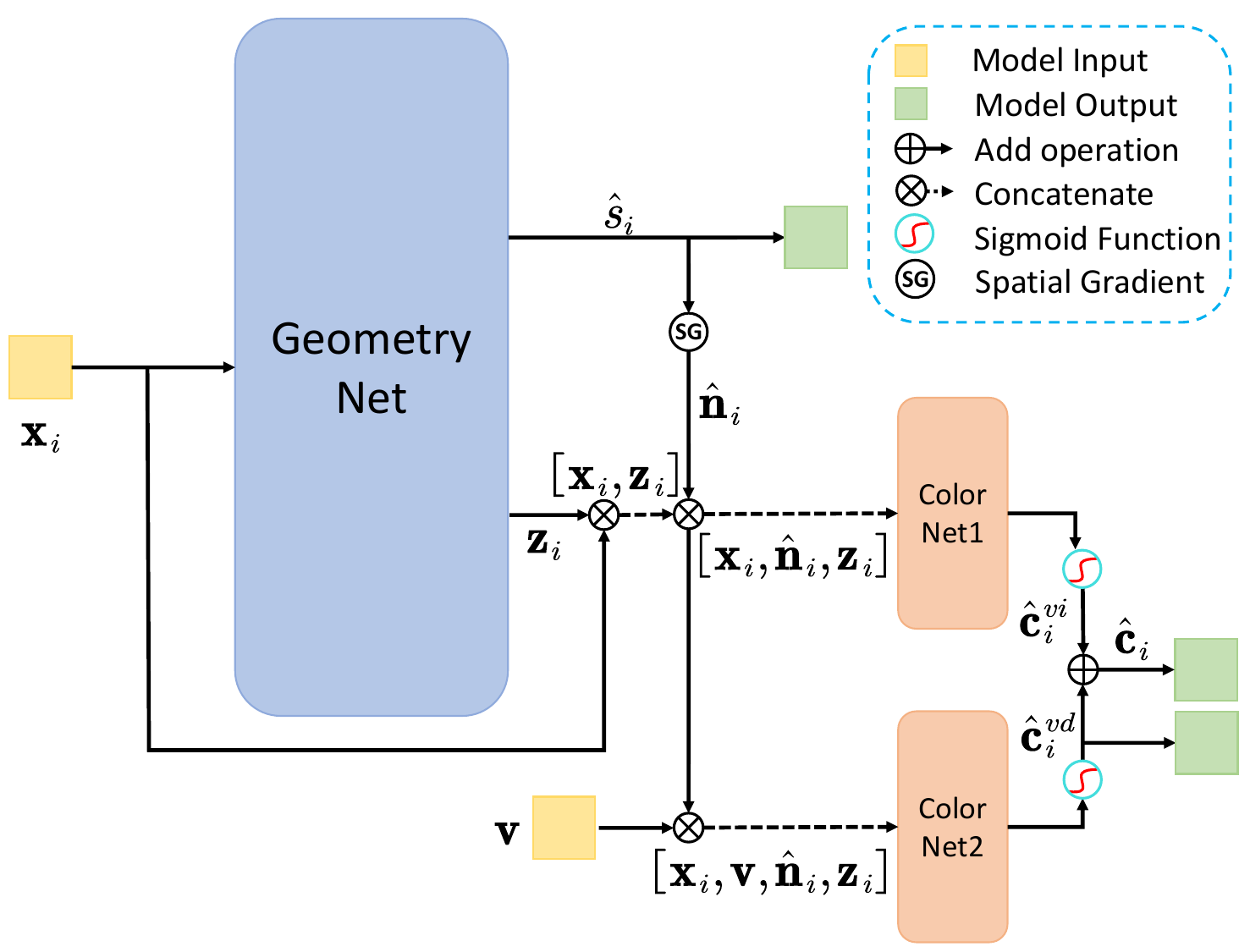}
\caption{Network architecture of our method.}

\label{network}
\vspace{-0.7cm}
\end{figure}

\subsection{Neural Surface Representation}

As shown in Fig. \ref{network}, the neural surface representation is composed of two kinds of MLPs: a geometric geometry network $f_g$ and two color networks $f_c$. The geometry network maps a 3D coordinate $\mathbf{x} \in \mathbb{R}^3 $ to a feature vector $\mathbf{z}$ and a signed distance function (SDF) value $\hat{s} \in \mathbb{R}$, which indicates the shortest distance to the closest geometry surface, and its sign shows whether the point is inside or outside the object. 

\begin{equation}
[\hat{s}, \mathbf{z}] = f_g(\mathbf{x}; \theta_g)
\end{equation}

\noindent where $\theta_g$ is the trainable parameters. The surface $S$ is defined as the set of $\mathbf{x}$ where the corresponding $\hat{s}$ is equal to 0. 

The color networks $f_c$ generate color $\hat{\mathbf{c}} \in \mathbb{R}^3$ from the input feature vectors.
We use two color networks to obtain view-dependent color $\hat{\mathbf{c}}^{vd}$ and view-independent color $\hat{\mathbf{c}}^{vi}$. 
This setting helps us to decompose color which we will discuss in Section \ref{pc_consistency}.

To recover the surface under the supervision of input RGB images, we render the colors of sampled rays. Taking render in the ray $\mathbf{r}$ as an example,  we feed $n$ sampling points $\mathbf{x}_{i} = \mathbf{o} + t_{i}\mathbf{v}$ along the ray into the geometric network $f_g$ to obtain the corresponding SDF values $\hat{s}_i$ and geometric feature $\mathbf{z}_i$. Here, $\mathbf{o}\in \mathbb{R}^3$ and $\mathbf{v}\in \mathbb{R}^3$ represents the camera position and ray direction with $\left| \mathbf{v}\right|=1$ and $t_i\geq0$, respectively. Subsequently, we concatenate $\mathbf{x}_i$, $\mathbf{v}$, $\mathbf{\hat{n}}_{i}$ and $\mathbf{z}_i$ into two feature vectors and obtain the corresponding view-dependent color $\mathbf{\hat{c}}_i^{vd}=f_c(\mathbf{x}_i, \mathbf{\hat{n}}_i, \mathbf{v}, \mathbf{z}_i; \theta_c)$ and view-independent color $\mathbf{\hat{c}}_i^{vi}=f_c(\mathbf{x}_i, \mathbf{\hat{n}}_i, \mathbf{z}_i; \theta_c)$. The normal vector $\mathbf{\hat{n}}_{i}\in \mathbb{R}^3$ is the analytical gradient of the corresponding SDF value $\hat{s}_i$, and the final color $\mathbf{\hat{c}}_i$ is obtained by adding two color components together:

\begin{equation}
[\hat{s}_i, \mathbf{z}_i] = f_{g}(\mathbf{x}_i), 
\quad
\hat{\mathbf{n}}_i = \frac{\partial \hat{s}_i}{\partial \mathbf{x}_i}
\end{equation}


\begin{equation}\label{color_sum}
\mathbf{\hat{c}}_i = \mathbf{\hat{c}}_i^{vd} + \mathbf{\hat{c}}_i^{vi}
\end{equation}

Following the volume rendering framework~\cite{mildenhall2021nerf, yariv2020multiview}, the color $\mathbf{C}(\mathbf{r})$ is accumulated along the ray.

\begin{equation} \label{color_render}
\hat{\mathbf{C}}(\mathbf{r}) = \sum_{i=1}^{n}T_i\alpha_i\hat{\mathbf{c}}_i
\end{equation}


\noindent where $T_i=\prod_{j=1}^{i-1}(1-\alpha_j)$ and $\alpha_i=1-exp(-\sigma_i\delta_i)$ denote the transmittance and alpha value, respectively. $\delta_i$ is the distance between neighbouring points, and $\sigma_i$ is the density value corresponding to $\mathbf{x}_i$. To improve the geometric representation and enhance the smoothness of the reconstructed surface, we compute density values $\sigma_i$ from $\hat{s}_i$~\cite{yariv2021volume, yu2022monosdf}:

\begin{equation}
\sigma_i(\hat{s}_i)=\left\{
\begin{array}{lcr}
\frac{1}{2\beta}exp(\frac{-\hat{s}_i}{\beta}), & & {if\ \hat{s}_i > 0}.\\
\frac{1}{\beta}(1 - \frac{1}{2}exp(\frac{\hat{s}_i}{\beta})), & & {if\ \hat{s}_i \leq 0}.
\end{array} \right.
\end{equation}


\noindent where $\beta$ is trainable. As $\beta$ approach 0, the sensitivity of $\sigma_i(\hat{s}_i)$ to $\hat{s}_i$ increases, contributing to edge reconstruction.


\subsection{Supervision Constraints}\label{spuervision_constrain}

\subsubsection{Color and Normal Constraints}\label{color_normal_constraints}

Since we have obtained each sample rays' rendering color $\hat{\mathbf{C}}(\mathbf{r})$, we can learn the weights of $f_g$ and $f_c$ by minimizing the difference between $\hat{\mathbf{C}}(\mathbf{r})$ and the given color $\mathbf{C}(\mathbf{r})$ :

\begin{equation}
\mathcal{L}_{rgb} = \sum_{\mathbf{r} \in \mathcal{R}}\left\|\hat{\mathbf{C}}(\mathbf{r}) - \mathbf{C}(\mathbf{r})\right\|_1
 \end{equation} 
 
\noindent where $\mathcal{R}$ represents the sampled rays in a batch.

Geometric properties like normal vectors $\hat{\mathbf{N}}(\mathbf{r})$ and depth $\hat{D}(\mathbf{r})$ can be rendered by accumulating sample points' features along the ray, similar to rendering colors. We utilize normal constraints to guide the training process: 

\begin{equation}\label{renderd_depth_and_normal}
\hat{D}(r) = \sum_{i=1}^{n}T_i\alpha_{i}t_i, \quad \hat{\mathbf{N}}(\mathbf{r}) = \sum_{i=1}^{n}T_i\alpha_{i}\hat{\mathbf{n}}_i  
\end{equation}

\begin{equation}\label{normal_loss}
\begin{split}
\mathcal{L}_{normal} = \frac{1}{|\mathcal{M}_r|} \sum_{\mathbf{r} \in \mathcal{M}_r}\left\|\hat{\mathbf{N}}(\mathbf{r}) - \bar{\mathbf{N}}(\mathbf{r})\right\|_1 \\
+ \left\|1-\hat{\mathbf{N}}(\mathbf{r})^{T}\bar{\mathbf{N}}(\mathbf{r})\right\|_1 
\end{split}
\end{equation}

\noindent where $\mathcal{M}_r$ denotes the ray mask. We will describe the details of $\mathcal{M}_r$ in Section \ref{mask_scheme}.

\subsubsection{Geometric Consistency Constraint}\label{gc_constraint}
\begin{figure}[htp]
\centering
\includegraphics[width=0.45\textwidth]{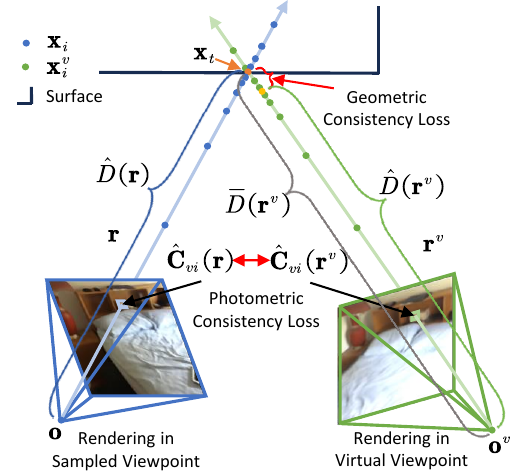}

\caption{Illustration of Consistency Constraints}
\vspace{-0.8cm}

\label{consistency_constraint}
\end{figure}

The proposed geometric consistency is based on the principle that geometric properties of the surface, such as depth or normal, should be consistent among different viewpoints in unobstructed regions. We utilize these consistencies, visualized in Fig. \ref{consistency_constraint}, to constrain the optimization process.

Specifically, for each sampled ray $\mathbf{r}$ passing through the current sampled pixel, we calculate the corresponding rendered depth $\hat{D}(\mathbf{r})$ and normal $\hat{\mathbf{N}}(\mathbf{r})$ (Eq. \ref{renderd_depth_and_normal}). Using render depth $\hat{D}(\mathbf{r})$, we compute the target point $\mathbf{x}_t$ (Eq. \ref{target_point}), which serves a similar purpose as the feature point in MVS but in 3D form. It is worth noting that this setting helps to avoid inaccuracies in feature extraction and matching errors. After that, we randomly generate a virtual viewpoint $\mathbf{o}^v$. Based on target point $\mathbf{x}_t$ and $\mathbf{o}^v$, we can calculate the virtual ray's direction $\mathbf{v}^v$. Consequently, we obtain a virtual ray $\mathbf{r}_v$ originating from $\mathbf{o}^v$ in direction $\mathbf{v}^v$, and virtual sampled points $\mathbf{x}^v_i = \mathbf{o}^v + t^v_i \mathbf{v}^v$, where $t^v_i \geq 0$, positioned along this ray.

\begin{equation}\label{target_point}
\mathbf{x}_t = \mathbf{o} + \hat{D}(\mathbf{r})\mathbf{v}, \quad \mathbf{v}^v = \frac{\mathbf{x}_t - \mathbf{o}^v}{ \left\|\mathbf{x}_t - \mathbf{o}^v\right\|_2 } 
\end{equation}

Using the volume rendering framework, we render the depth $\hat{D}(\mathbf{r}_v)$ and normal $\hat{\mathbf{N}}(\mathbf{r}_v)$ of $\mathbf{r}_v$. Due to the geometric consistency between the depth of both rays, we propose a novel optimization target:



\begin{equation}\label{gc_loss}
\mathcal{L}_{gc} = \frac{1}{2|\mathcal{M}_v|}\sum_{\mathbf{r}_v \in \mathcal{M}_v}|\hat{D}(\mathbf{r}_v)-\bar{D}(\mathbf{r}_v)|^2
\end{equation}

\noindent where $\bar{D}(\mathbf{r}_v) = \left\|\mathbf{x}_t - \mathbf{o}^v\right\|_2$, and $\mathcal{M}_v$ denotes the mask for valid sample rays but failed in multi-view normal consistency. The details of $\mathcal{M}_v$ will be described in Section \ref{mask_scheme}.

\subsubsection{Photometric Consistency Constraint}
\label{pc_consistency}
Similar to the geometric consistency across views, the appearance of the scene also exhibits consistency. Due to changes in illumination or material properties, colors may appear different from various viewpoints. Inspired by~\cite{tang2022nerf2mesh}, we decompose the render color of each sample point. Concretely, we leverage two color networks to predict view-dependent color $hat{\mathbf{c}}^{vd}_i$ and view-independent color $\hat{\mathbf{c}}^{vi}_i$, as shown in Fig. \ref{network}. The final rendering color $\hat{\mathbf{c}}_i$ is obtained by summing these two terms, as shown in Eq. \ref{color_sum}.


The view-independent colors $\hat{\mathbf{C}}_{vi}$ of two kinds of rays are accumulated (Eq. \ref{color_render}). We propose an additional photometric consistency:

\begin{equation}\label{pc_loss}
\mathcal{L}_{pc} = \frac{1}{|\mathcal{M}_r|} \sum_{\mathbf{r} \in \mathcal{M}_r}\left\| \hat{\mathbf{C}}_{vi}(\mathbf{r}) - \hat{\mathbf{C}}_{vi}(\mathbf{r}_v)\right\|_1
\end{equation}





\subsection{Mask Scheme}\label{mask_scheme}
In this section, we will introduce our mask scheme applied in the second training stage and utilize the AND operation to combine masks into $\mathcal{M}_v$ and $\mathcal{M}_r$. It is worth noting that we present the valid rays as 1 in all of our masks. 

\subsubsection{Sample Mask}

To enforce multi-view consistency, a virtual viewpoint $\mathbf{o}^v$ is randomly generated for each sampled ray. However, this setting may result in $\mathbf{o}^v$ being positioned outside the scene or inside objects. To address this issue, we propose a sample mask $\mathcal{M}_s$ to select valid virtual viewpoints. Specifically, we primarily utilize the SDF value in $\mathbf{o}^v$. Our reconstruction starts with a sphere that encloses all the given camera poses. As the training progresses, this sphere gradually approaches our target. Consequently, the outside part can be considered as the interior of the object, which means that if $\mathbf{o}^v$ is valid, we will get a positive corresponding SDF value $\hat{s}(\mathbf{o}_v)$. Our sample mask is as follows:

\begin{equation}
\mathcal{M}_s=\left\{
\begin{array}{lcr}
1, & & {if\ \hat{s}(\mathbf{o}_v)>0}.\\
0, & & otherwise.
\end{array} \right.
\end{equation}

\subsubsection{Occlusion Mask}\label{occlusion_mask}
To address the problem of errors in depth consistency caused by occlusion along both rays, we propose an occlusion mask $\mathcal{M}_o$. Following the sampling algorithm in~\cite{yariv2021volume}, our sampled points are concentrated near the surfaces where the rays pass through. Hence, we can identify the presence of occlusion by analyzing the sign change in the SDF values associated with the sampling points along the ray.

\[
\mathcal{M}_{o}^s=\left\{
\begin{array}{lrr}
1, & {if\ \left\|\mathit{diff}(\mathit{sgn}(\mathbf{\hat{s}}))\right\|_{1}\leq 2}.\\
0, & otherwise. 
\end{array} \right.
\]

\[
\mathcal{M}_{o}^v=\left\{
\begin{array}{lrr}
1, & {if\ \left\|\mathit{diff}(\mathit{sgn}(\mathbf{\hat{s}}^v))\right\|_{1}\leq 2}.\\
0, & otherwise. 
\end{array} \right.
\]

\begin{equation}
\mathcal{M}_o = \mathcal{M}_{o}^s \ \& \ \mathcal{M}_{o}^v
\end{equation}

\noindent where $\mathit{diff}(\cdot)$ computes the $n$-th forward difference along the given vector's dimension, and $\mathit{sgn}(\cdot)$ is the sign function. $\mathbf{\hat{s}}$ and $\mathcal{M}_{o}^s$ denote the vector of SDF values along the sample ray and the occlusion mask of this ray. Similarly, $\hat{\mathbf{s}}^v$ and $\mathcal{M}_{o}^v$ represent the corresponding values for the virtual rays. Finally, the final occlusion mask $\mathcal{M}_o$ is obtained through the AND operation between $\mathcal{M}_o^s$ and $\mathcal{M}_o^v$.


\subsubsection{Adaptive Check Mask}\label{adaptive_check_mask}
As described in Section \ref{gc_constraint}, high-quality reconstruction conforms to geometric consistency in multi-views. Therefore, we utilize the consistency of the rendered normal in multi-views as an adaptive check. Specifically, we use the normal Cosine Similarity (Eq. \ref{cosine_similarity}) to compute the difference between the sample ray's render normal $\hat{\mathbf{N}}(\mathbf{r})$ and the virtual ray's render normal $\hat{\mathbf{N}}(\mathbf{r}_v)$. We compare this difference to a certain threshold value $\epsilon$. Rays with significant differences are identified by $\mathcal{M}_a$:

\begin{equation}\label{cosine_similarity}
cos(\hat{\mathbf{N}}(\mathbf{r}), \hat{\mathbf{N}}(\mathbf{r}_v)) = \frac{\hat{\mathbf{N}}(\mathbf{r}) \cdot \hat{\mathbf{N}}(\mathbf{r}_v)}{\left\| \hat{\mathbf{N}}(\mathbf{r}) \right\|_2 \left\| \hat{\mathbf{N}}(\mathbf{r}_v) \right\|_2}
\end{equation}

\begin{equation}
\mathcal{M}_a=\left\{
\begin{array}{lcr}
1, & & {if\ cos(\hat{\mathbf{N}}(\mathbf{r}), \hat{\mathbf{N}}(\mathbf{r}_v)) < \epsilon}.\\
0, & & otherwise.
\end{array} \right.
\end{equation}

\subsubsection{Mask integration}
To better utilize the estimated normal cues and multi-view consistency, we organize the sample ray masks by AND operations:

\begin{equation}
\begin{split}
\mathcal{M}_v =& \,\, \mathcal{M}_s \ \& \ \mathcal{M}_o \ \& \ \mathcal{M}_a \\
\mathcal{M}_r =& \,\, \mathcal{M}_s \ \& \ \mathcal{M}_o \ \& \ (1 - \mathcal{M}_a) 
\end{split}
\end{equation}

Rays selected by $\mathcal{M}_v$ have valid virtual viewpoints and no occlusion issues but fail to check in multi-view normal consistency. We put the geometric consistency constraint on these rays' training process.  $\mathcal{M}_r$ chooses rays that conform to the normal consistency check. In those rays, predicted normal cues contribute to the reconstruction and we continue to apply them. In addition, we incorporate photometric consistency to further improve the quality.


\section{EXPERIMENTS}\label{experiments}

\subsection{Implementation Detail}
We implement our method with PyTorch and the network training is performed on one NVIDIA RTX 3090 GPU. The normal priors in our method are predicted by Omnidata model~\cite{eftekhar2021omnidata}. Each batch consists of 1024 sampled rays and training the network over 200k iterations. We first optimize the model directly guided with normal priors and RGB images over 25k iterations. In the second stage, in addition to color constraint, our model is trained under mask-guided normal and geometric consistency constraints until 75k iterations. After that, we add photometric consistency into our optimization target. Following the optimization, we discrete our implicit function into voxel grids with a resolution of 512 and extract the mesh using Marching Cubes~\cite{lorensen1998marching}.

\subsection{Experimental Settings}

\subsubsection{Datasets}
Since our method primarily focuses on indoor scenes, we conduct quantitative evaluations on the Replica dataset~\cite{straub2019replica} and the ScanNet dataset~\cite{dai2017scannet}. Replica consists of high-quality reconstructions of various indoor spaces. Each scene in Replica offers clean dense geometry and high-resolution images captured from multiple viewpoints. ScanNet is an RGB-D video dataset that comprises over 1500 indoor scenes with 2.5 million views. It is annotated with ground-truth camera poses, surface reconstructions, and instance-level semantic segmentations.

\subsubsection{Baselines}
We conduct a comparative analysis of our method with other methods. (1) COLMAP~\cite{schonberger2016pixelwise}: Traditional MVS reconstruction method, using screened Poisson Surface reconstruction (sPSR) to reconstruct mesh from point clouds. (2) NeuralRecon~\cite{sun2021neuralrecon}: A learning-based TSDF fusion module. (3) MonoSDF(MLP version)~\cite{yu2022monosdf}: Implicit method using predicted normal and depth priors directly. (4) NeuRIS~\cite{wang2022neuris}: Implicit method adaptive using normal priors. 

\subsubsection{Evaluation Metrics}
To evaluate the quality of scene representation, following~\cite{guo2022neural, murez2020atlas, yu2022monosdf} , we mainly report \textit{Accuracy} , \textit{Completeness}, \textit{Chamfer Distance}, \textit{Precision}, \textit{Recall} and \textit{F-score} with the threshold of 0.05 meter. We further report \textit{Normal Consistency} measure~\cite{yu2022monosdf} to better evaluate reconstructions under the synthetic dataset.

\begin{table}[ht]
\vspace{-0.2cm}
\centering
\caption{quantitative results on the ScanNet dataset}
\vspace{-0.3cm}

\setlength{\tabcolsep}{0.8mm}{
\resizebox{.48\textwidth}{!}{
\begin{tabular}{ccccccc}
\hline
\textbf{Method} & \textbf{Acc.$\downarrow$} & \textbf{Comp.$\downarrow$} & \textbf{Chamfer-$L_1$ $\downarrow$} & \textbf{Prec.$\uparrow$} & \textbf{Recall$\uparrow$} & \textbf{F-score$\uparrow$} \\

\hline
\textbf{COLMAP} & 0.047 & 0.235 & 0.141 & 0.711 & 0.441 & 0.537  \\
\textbf{NeuralRecon}  & 0.044 & 0.123 & 0.084 & 0.741 & 0.502 & 0.595\\
\textbf{NeuRIS} & 0.050 & 0.049 & 0.050 & 0.717 & 0.669 & 0.692  \\
\textbf{MonoSDF(MLP)} & \textbf{0.035} & 0.048 & 0.042 & 0.799 & 0.681 & 0.733 \\
\hline

\textbf{Ours}	& 0.036 & \textbf{0.045} & \textbf{0.040} & \textbf{0.837} & \textbf{0.734} & \textbf{0.780} \\

\hline
\end{tabular}
}}
\vspace{-0.5cm}
\label{scannet_table}
\end{table}

\subsection{Results in Realistic Dataset}
\begin{figure*}[ht]
\centering
\vspace{-0.4cm}
\includegraphics[width=0.9\textwidth]{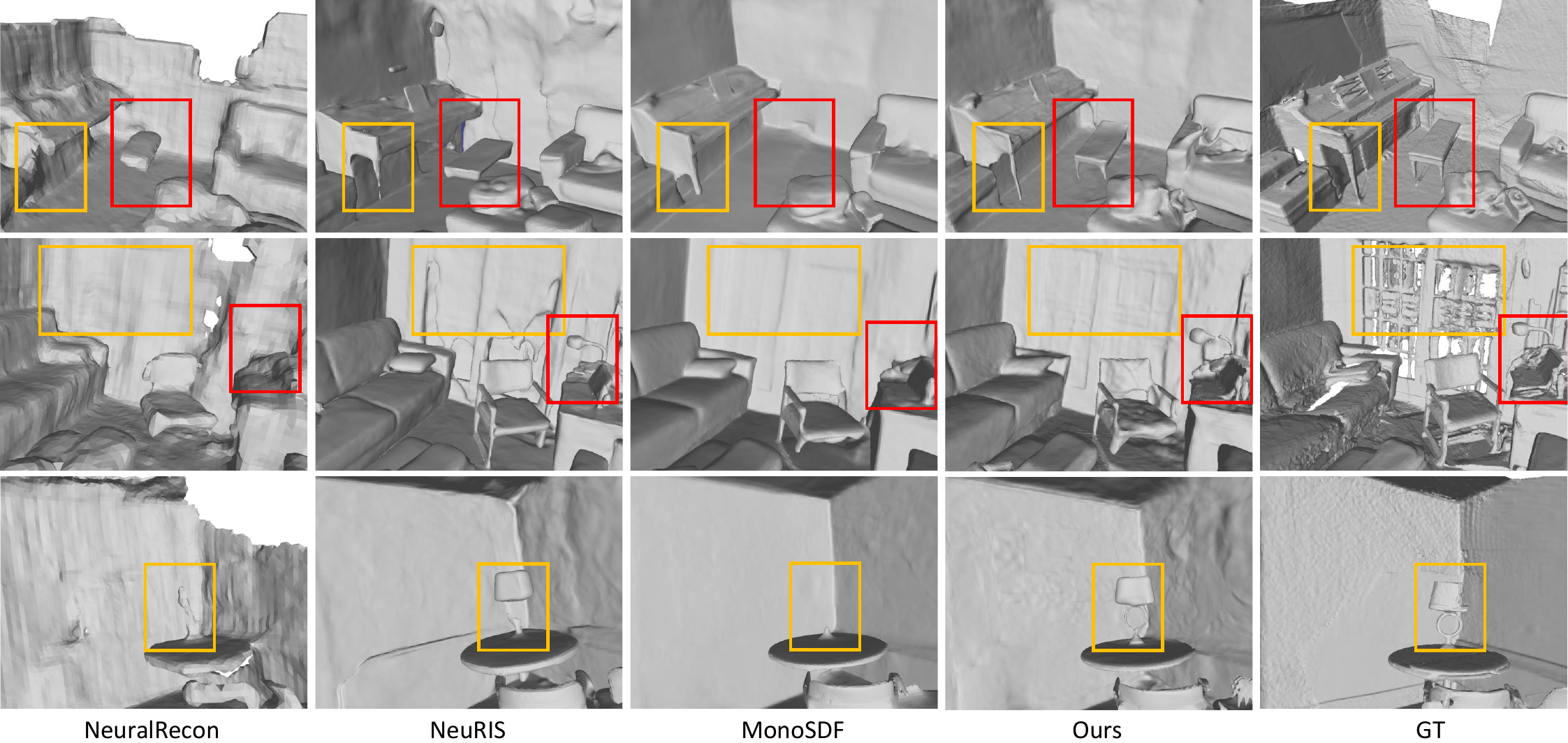}
\caption{Qualitative comparison of indoor 3D surface reconstruction on the ScanNet dataset.}

\label{scannet_visual}
\vspace{-0.6cm}
\end{figure*}

We conducted experiments using ScanNet dataset, which provides real-world data. We selected four scenarios and chose every 10th image from the original sets (about 2k-4k images). After the whole training process, we saved the reconstructed mesh and evaluated the final trained model. In Fig. \ref{scannet_visual}, we compare our generated mesh with the original reconstructions from baselines. It shows that ours has more geometric detail and fixes missing caused by inaccurate estimation of normals. Additionally, we quantitatively evaluated our method compared to others in Table \ref{scannet_table}. It can be seen that our method achieves more accurate results.

COLMAP~\cite{schonberger2016pixelwise} exhibits limitations in low-texture regions. NeuralRecon~\cite{sun2021neuralrecon} relies on TSDF values for supervision and suffers from limitations in unseen scenarios. The performance of MonoSDF~\cite{yu2022monosdf} is dependent on the quality of predicted geometric cues, which may lead to inaccuracy. NeuRIS~\cite{wang2022neuris} incorporates normal priors adaptively but is susceptible to noise in real-world datasets, resulting in artifacts in reconstructions.

\subsection{Results in Synthetic Datasets}
We conducted a quantitative evaluation of five scenarios from the Replica and averaged the results for comparison. It is worth noting that Replica is a synthetic dataset with little noise, which makes the predicted cues have minimal errors. To demonstrate the effectiveness of our approach in utilizing normal priors, we compare ours with modified MonoSDF* that directly utilizes normal priors and images for supervision. We conducted a comparison in Table \ref{replica_table}, and it indicates that our method achieves more accurate results, though the majority of the normal vectors are accurate.

\begin{table}[ht]
\vspace{-0.2cm}
\centering
\caption{quantitative comparison on the replica dataset
  }
  
\vspace{-0.3cm}
\setlength{\tabcolsep}{1.5mm}{
\resizebox{.48\textwidth}{!}{
\begin{tabular}{cccccccccc}

\hline
{} & \multicolumn{2}{c}{\textbf{Chamfer-$L_1$ $\downarrow$}} & \multicolumn{2}{c}{\textbf{F-score $\uparrow$}}  & \multicolumn{2}{c}{\textbf{Normal. C $\uparrow$}} \\

\textbf{Scan} & \textbf{monosdf*}& \textbf{Ours} &\textbf{monosdf*} & \textbf{Ours} &\textbf{monosdf*} & \textbf{Ours} \\
\hline
scan1 & 4.104 & 2.593 & 76.945  & 89.589 & 88.133 & 91.619 \\

scan2 & 2.415 & 2.996 & 92.316  & 89.045 & 93.945 & 94.304 \\

scan3 & 6.292 & 3.718 & 85.153 & 85.843 & 93.851 & 93.376 \\

scan4 & 3.119 & 2.562 & 86.367 & 92.031 & 91.290 & 92.329 \\

scan5 & 4.627 & 3.900 & 84.976 & 85.604 & 92.244 & 93.833 \\

\hline
mean & 4.111 & \textbf{3.154} & 85.151 & \textbf{88.423} & 91.893 & \textbf{93.029} \\
\hline

\end{tabular}
}}
\label{replica_table}
\vspace{-0.7cm}
\end{table}

\subsection{Ablation Studies}
We conducted an ablation study to analyze the impact of consistency constraints and the adaptive mask in our method. To do this, we removed each setting from our framework and evaluated the results. We randomly selected three scenarios from the ScanNet and averaged the final evaluated metrics to perform the ablation experiment. The comprehensive results of the ablation study can be found in Table \ref{ablation_table}.

\begin{table}[h]
\vspace{-0.2cm}
\caption{Quantitative results of ablation studies}
\label{ablation_table}
\centering

\vspace{-0.3cm}

\setlength{\tabcolsep}{0.2mm}{
\resizebox{.48\textwidth}{!}{
\begin{tabular}{cccccccccc}
\hline


\textbf{Norm}.	& \textbf{Mask} 	  &\textbf{Geo.}	  &\textbf{Pho.}   &\textbf{Acc.$\downarrow$}	&\textbf{Comp.$\downarrow$}	&\textbf{Chamfer-$L_1 \downarrow$}	&\textbf{Prec.$\uparrow$}	&\textbf{Recall$\uparrow$}	&\textbf{F-score$\uparrow$} \\

\hline
\checkmark  &              &             &             & 3.539  & 5.042  & 4.290  & 80.051  & 65.162  & 71.752 \\
\checkmark  &  \checkmark  & \checkmark  &             & 3.519  & 4.990 & 4.254  & 80.392  & 65.652  & 72.204 \\

\checkmark  &  \checkmark  &	         & \checkmark  & 3.526  & 4.864  & 4.195  & 80.834  & 66.832  & 73.094\\

\checkmark  &  		   & \checkmark  & \checkmark  & 3.546  & 4.949  & 4.248  & 80.428  & 65.561  & 72.170 \\

\checkmark  &  \checkmark  & \checkmark  & \checkmark  & \textbf{3.247}  & \textbf{4.667}  &\textbf{3.957}   & \textbf{84.079}  & \textbf{69.696}  & \textbf{76.149} \\

\hline
\end{tabular}
}}
\vspace{-0.3cm}
\end{table}











\subsubsection{Effectiveness of Geometric Consistency} 
In regions that failed in normal consistency checks, we add geometric constraints into the training process to improve geometric detail reconstruction. In Table \ref{ablation_table}, geometric constraints improved quantitative results in terms of F-score and Chamfer-$L_1$ metrics, compared with directly utilizing normal priors. This demonstrates the effectiveness of the geometric consistency.

\subsubsection{Effectiveness of Photometric Consistency}
We utilize photometric consistency for further improvement in areas that conform to normal consistency. Cause the majority of normal priors are accurate, most optimization will constrained by photometric consistency rather than geometric consistency. Therefore, this setting is more effective than only using geometric consistency, as shown in Table \ref{ablation_table}

\subsubsection{Effectiveness of Adaptive Mask}
To enhance the effectiveness of constraints, we employ an adaptive mask for selection. As shown in Table \ref{ablation_table}, applying all constraints without selection has limited improvement, due to the conflict between the inaccurate normal priors constraint and consistency constraints.

\begin{table}[h]
\vspace{-0.2cm}
\centering
\caption{the effectiveness of color decomposition }

\vspace{-0.3cm}
\setlength{\tabcolsep}{0.7mm}{
\resizebox{.48\textwidth}{!}{
\begin{tabular}{ccccccc}
\hline
{} & \textbf{Acc.$\downarrow$}	&\textbf{Comp.$\downarrow$}	&\textbf{Chamfer-$L_1 \downarrow$}	&\textbf{Prec.$\uparrow$}	&\textbf{Recall$\uparrow$}	&\textbf{F-score$\uparrow$} \\
\hline

full(w/o CD.) & 3.576 & 4.940 & 4.258 & 80.323 & 66.094 & 72.445 \\
full & \textbf{3.247}  & \textbf{4.667}  &\textbf{3.957}   & \textbf{84.079}  & \textbf{69.696}  & \textbf{76.149} \\
\hline
\end{tabular}
}}
\label{color decomposed }
\vspace{-0.3cm}
\end{table}

Table \ref{ablation_table} shows that all of the settings contribute to enhancing reconstruction results, and lead to higher-quality reconstructions when employed in conjunction. Furthermore, to demonstrate the effectiveness of color decomposition, we conducted a comparative experiment on color. As shown in Table \ref{color decomposed }, color decomposition also contributes to reconstruction.

\section{CONCLUSION}\label{conclusion}

In this paper, we propose a 3D indoor scene reconstruction method using a two-stage training process. We decompose the color by perspective dependency and utilize two novel consistency constraints in both geometry and photography to improve the reconstruction quality. Besides, we introduce an essential mask to effectively select constraints in the training process. The experiments show that our approach achieves more geometric detail and high-quality reconstruction.



\bibliographystyle{IEEEtran.bst}
\bibliography{ref.bib}

\end{document}